\documentclass[final]{OAGM}
\OAGMarXiv{1404.3538}
\pdfoutput=1 
\usepackage{graphicx}
\usepackage{epstopdf}
\usepackage{multirow}
\usepackage{amsmath}
\usepackage[bf,sf]{subfigure}

\title{Indoor Activity Detection and Recognition for Sport Games Analysis}
\author{Georg Waltner \and Thomas Mauthner \and Horst Bischof\\
  Institute for Computer Graphics and Vision, University of Graz, Austria}
\begin{document}
\maketitle
\begin{abstract}
Activity recognition in sport is an attractive field for computer vision research. Game, player and team analysis are of great interest and research topics within this field emerge with the goal of automated analysis. The very specific underlying rules of sports can be used as prior knowledge for the recognition task and present a constrained environment for evaluation. This paper describes recognition of single player activities in sport with special emphasis on volleyball. Starting from a per-frame player-centered activity recognition, we incorporate geometry and contextual information via an activity context descriptor that collects information about all player's activities over a certain timespan relative to the investigated player. The benefit of this context information on single player activity recognition is evaluated on our new real-life dataset presenting a total amount of almost 36k annotated frames containing 7 activity classes within 6 videos of professional volleyball games. Our incorporation of the contextual information improves the average player-centered classification performance of 77.56\% by up to 18.35\% on specific classes, proving that spatio-temporal context is an important clue for activity recognition.
\end{abstract}

\section{Introduction}
\label{sec:introduction}
Originally, activity recognition focused on evaluation of isolated single person behavior. While there has been effort in this recognition task for many years, the focus of information extraction for description has been extended from motion and shape features to spatio-temporal context information within the last years. State-of-the-art research was focused predominantly on the level of individuals, with less emphasis on group aspects. However, behavior of individuals is influenced by their surroundings, for example by interaction with other individuals and objects, or by the natural scene boundaries. One could summarize these as local (spatial and temporal) scene context, which is influencing the current and future behavior of an individual. Especially team sports like volleyball are structured concerning the role of individual players or positions on the court.\\
This paper presents a new method for activity recognition in sport with emphasis on volleyball, as a representative for indoor sport. Assuming known geometry of the scene by the court as a common ground plane between the videos, we start with a per-frame investigation and description of sport-specific single player activities by applying standard appearance (Histograms of Gradients (HOG) \cite{dalal05}) and motion (Histograms of Oriented Flow  (HOF) \cite{laptev08}) features which have a reputation for working well if object regions are known. In addition we exploit location information of an observed player (Real World Player Coordinates (RWPC)) along with occupancy probabilities of all players on the ground-plane (Spatial Context (SC) descriptor). Upon these features activities of individual players are classified via a Support Vector Machine (SVM). Subsequently, the single player classification results are embedded as context information by introducing the Activity Context (AC) descriptor, for description of activity probabilities on the court.
\section{Related Work}
\label{sec:related}
In the beginning of action/activity recognition single persons were examined separately. Different types of descriptors emerged based on motion, shape, key-poses, body part models or keypoint trajectories. The first descriptors used keypoint detectors for collection of simple features like corners or edges. A popular example is the Harris corner detector proposed by \cite{harris88}, that was later extended to 3D in \cite{laptev03}. Then, more complex descriptors introducing shape and motion features were proposed. In \cite{davis97}, activities like "sitting down" or "waving" as well as aerobic exercises were examined by using motion-energy images (MEI) and motion-history images (MHI). The widely used HOG descriptor \cite{dalal05} models shape and is often used together with the HOF descriptor \cite{laptev08} for motion characterization. Popular examples of keypoint descriptors are the scale-invariant feature transform (SIFT) proposed in \cite{lowe04} and the 3D SIFT descriptor \cite{scovanner07}. Spatio-temporal interest points (STIP) as introduced in \cite{laptev03} exploit both dimensions simultaneously. After recognition of single persons, the recognition has been extended to groups or even crowds. Naturally the relationship between these persons can give clues about single person's actions, which is the reason for introduction of context descriptors for both spatial and temporal dimension.\\
In \cite{lan12}, activities of groups in surveillance videos were examined by describing the activity of an individual person as well as the behavior of other persons nearby. This is related to the presented approach where first individual players are analyzed and then the analysis is combined over all players on the field. In \cite{choi11}, \textit{collective activities} like "queuing", "crossing", "waiting" or "talking" were recognized by building a spatio-temporal context descriptor based on positioning and motion features around every person in the frame. Other than in this paper, the spatial binning was not calculated globally, but circular with the examined person in the center. In \cite{zhu13}, individual activities in a scene were connected to create an activity context. With the segmented motions (continuous motions divided into action segments) in a video, these segments are set in context between themselves. Action segments that are related to each other in space and time are grouped together into activity sets. The combination of spatial and temporal context helps distinguishing between activities.
\\\textbf{Activity Recognition in Sport.} An approach for field hockey was presented in \cite{bialkowski13}, where a hockey field was recorded by eight HD cameras and players of both teams extracted by background subtraction and color models. Team activities were expressed by position context with occupancy maps and elliptical team centroids.
For activity recognition in basketball games trajectory features of the players were deployed. After a coach designs a strategic code-book with different complex defense or attack activities involving several players, the tracking results (trajectories) are compared to the templates in the code-book in \cite{perse08} and \cite{perse06}. Similar to \cite{bialkowski13}, the work of \cite{gade13} uses occupancy maps to recognize the type of sport within a sports arena.	Player positions, represented as Gaussian distributions, are combined over time into heatmaps that correlate to a individual sport type (badminton, basketball, handball, soccer, volleyball and miscellaneous). In contrast to our method, all above mentioned sport activity recognition systems only recognize team activities but not player activities.
\section{Proposed Method}
\begin{figure}[htbp!]
  \begin{center}
		\subfigure[]
		{
			\includegraphics[width=0.115\textwidth]{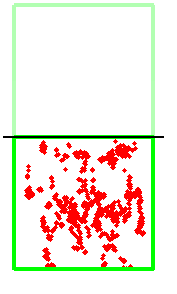}
		}
		\subfigure[]
		{
			\includegraphics[width=0.115\textwidth]{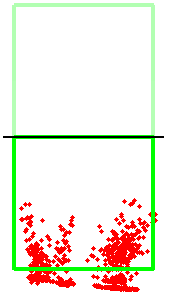}
		}
    \subfigure[]
		{
      \includegraphics[width=0.115\textwidth]{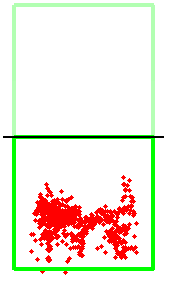}
    }
    \subfigure[]
		{
      \includegraphics[width=0.115\textwidth]{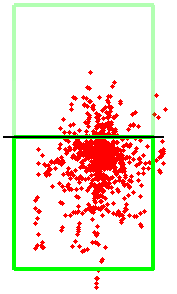}
    }
		\subfigure[]
		{
			\includegraphics[width=0.115\textwidth]{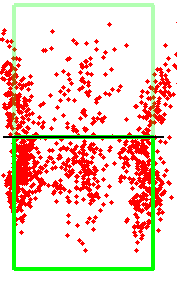}
		}
		\subfigure[]
		{
			\includegraphics[width=0.115\textwidth]{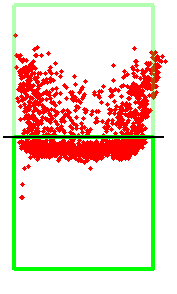}
		}
		\subfigure[]
		{
			\includegraphics[width=0.115\textwidth]{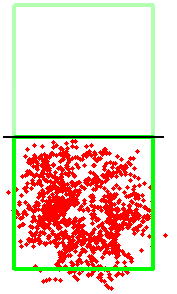}
		}
  \end{center}
  \caption{\small\textit{Positions of players during activities (top view): (a) "Stand", (b) "Service", (c) "Reception", (d) "Setting", (e) "Attack", (f) "Block", (g) "Defense/Move". Courts are marked in green, positions are shown as red dots. The black line indicates the net. Due to the planar homography, jumping players seem to be in the opposite court.}}
  \label{fig:action_locations}
\end{figure}

\begin{figure}[htbp!]
  \begin{center}
    \subfigure[Frame with grid, exemplary points $\boldsymbol{x}$ (1 to 6) and scaled rectangles $\Omega_{\boldsymbol{x}}$]
		{
      \includegraphics[width=0.22\textwidth]{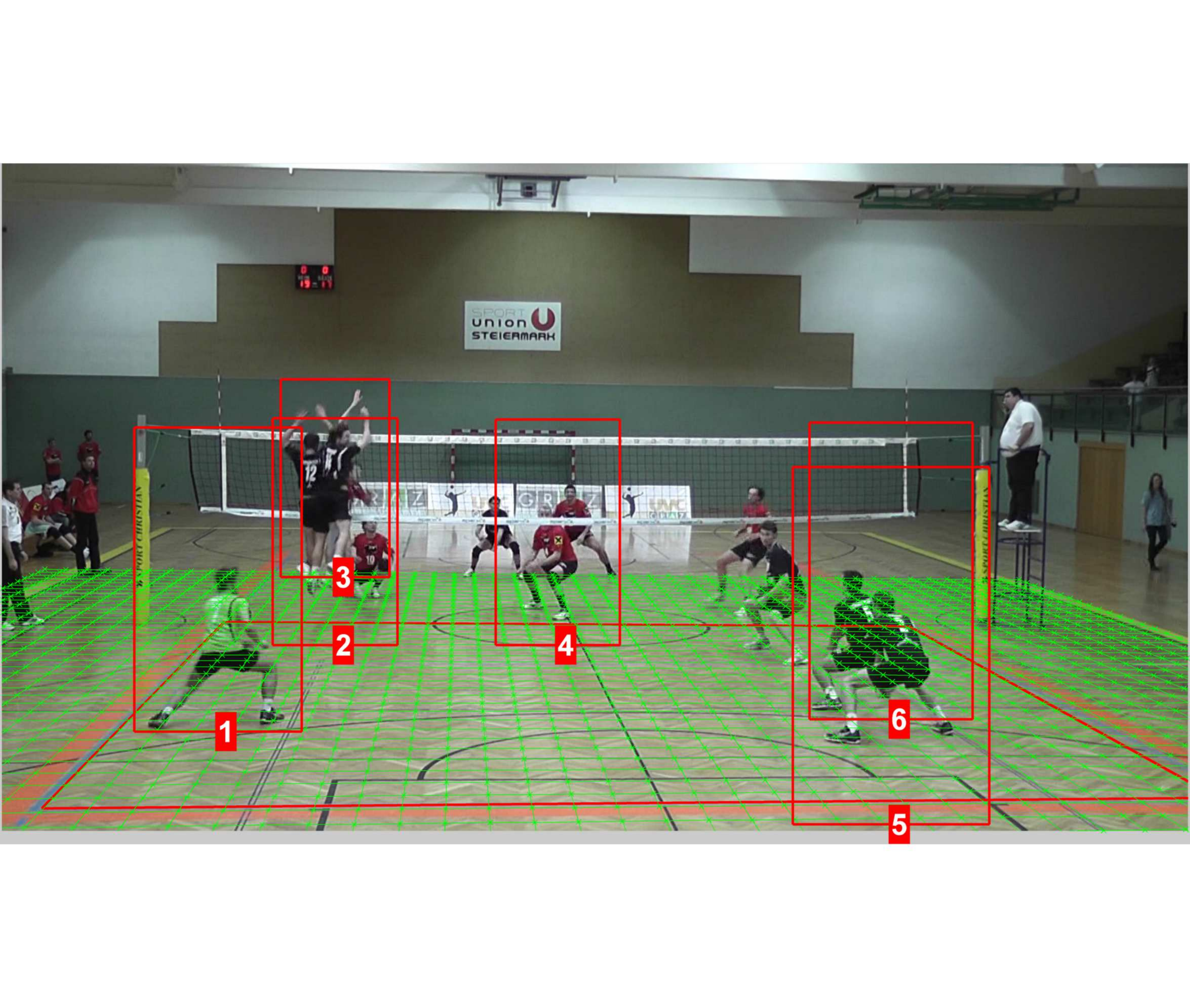}
    }
    \subfigure[Player probability image $P_{player}(\boldsymbol{x})$]
		{
      \includegraphics[width=0.22\textwidth]{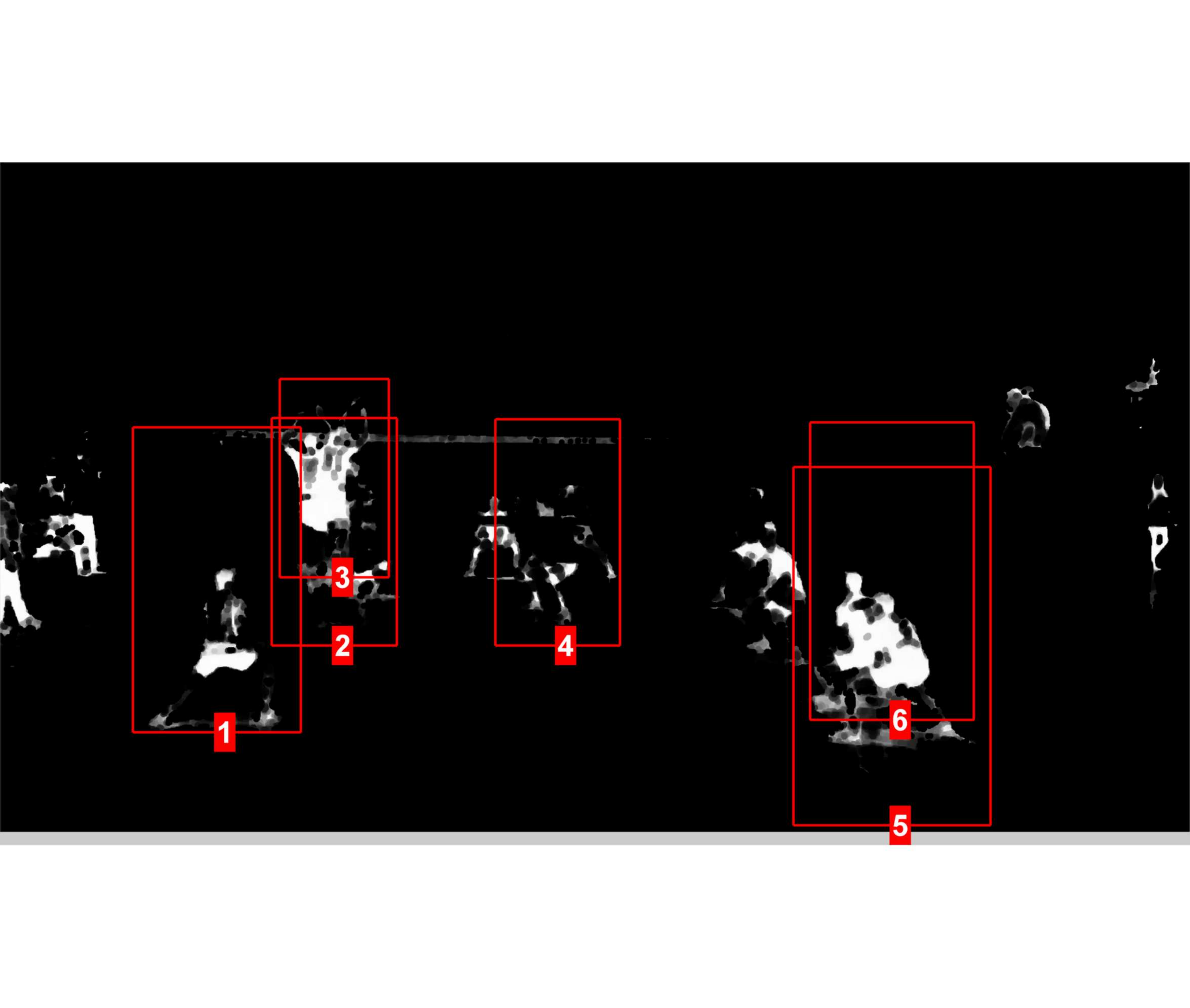}
    }
		\subfigure[Top view of the grid and transformed points $\boldsymbol{\tilde{x}}$]
		{
			\includegraphics[width=0.22\textwidth]{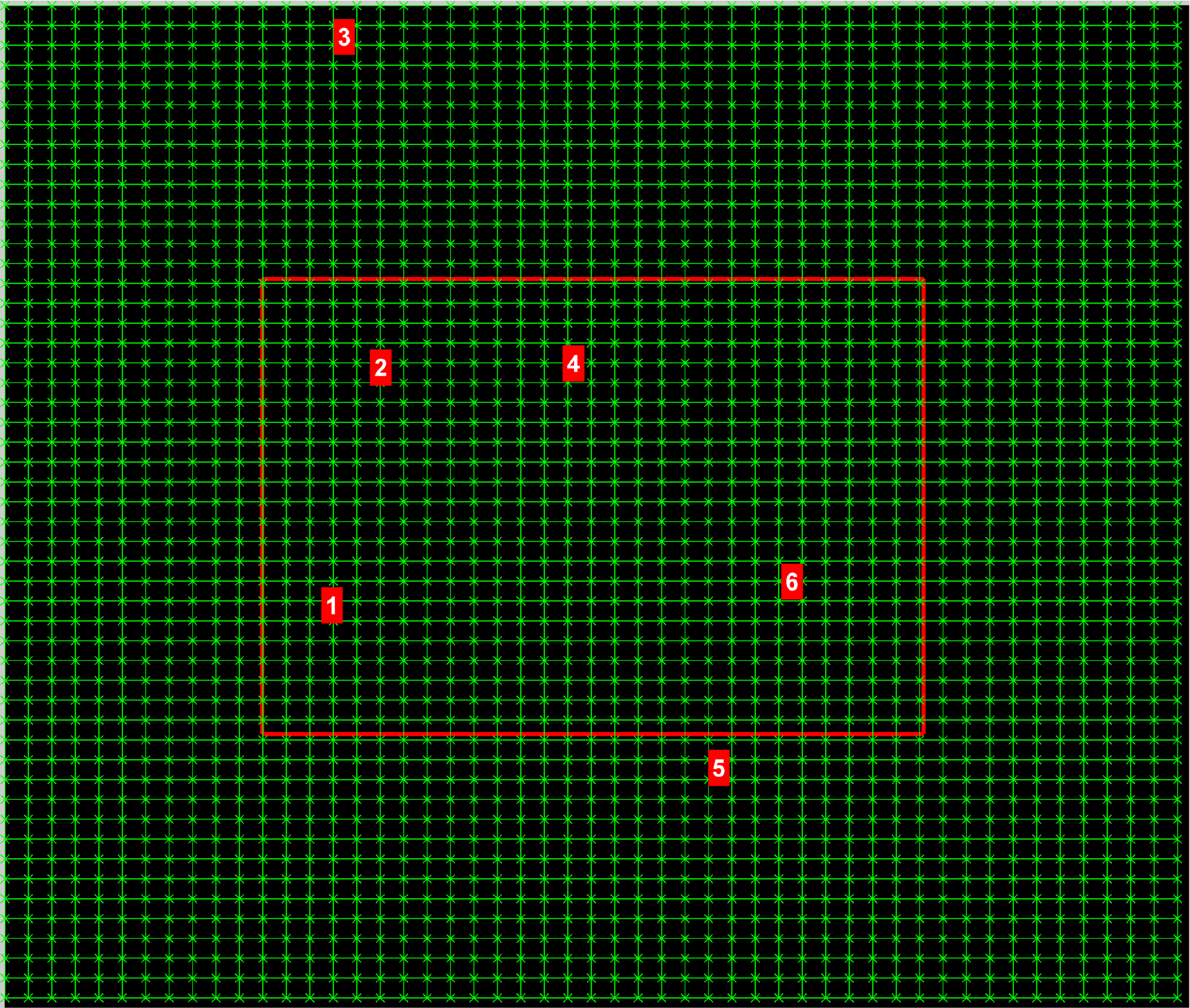}
		}
		\subfigure[Final SC descriptor, binned into 15x20 bins]
		{
			\includegraphics[width=0.22\textwidth]{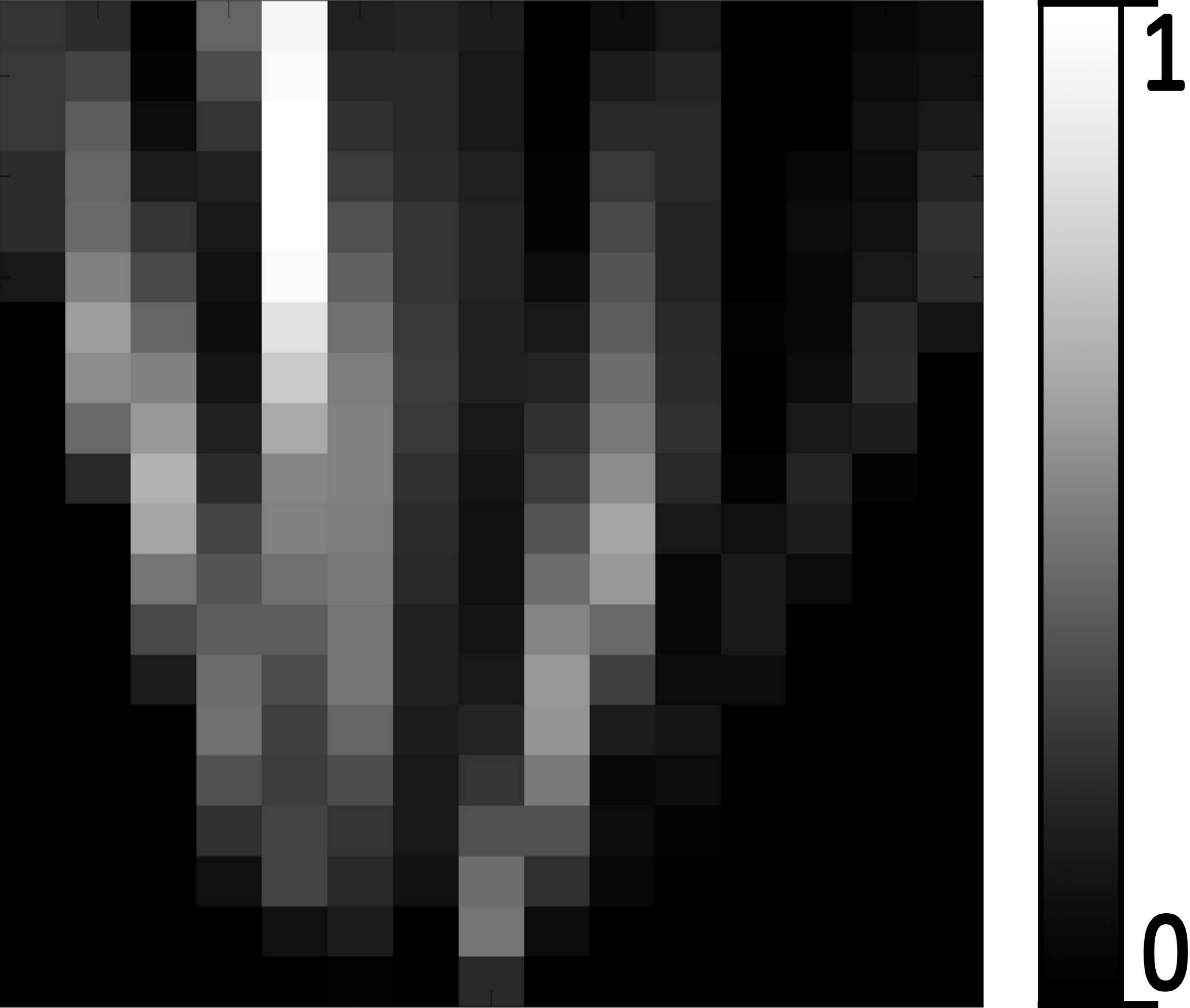}
		}
  \end{center}
  \caption{\small\textit{SC descriptor calculation: After calculating the player probabilities $P_{player}$ for the current frame $F$, for every point on the grid a corresponding rectangle $\Omega$ and the filled area percentage $\Pi$ is computed. The resulting player occupancy probabilities (ranging from 0 to 1) are binned for dimension reduction. Note: Players closer to the camera (rectangle 5) also fill out rectangles further away (6), such blurring the descriptor. Also, players at the net (partly) fill multiple rectangles around them, as the step size is getting smaller with the distance from the camera (2,3).}}
  \label{fig:context}
\end{figure}

Following the diction in \cite{turaga08}, our method for activity recognition is based on multiple steps: First input videos are calibrated and preprocessed, then player specific features (HOG, HOF, RWPC) are extracted. These features are combined with spatial context features (SC) and an activity classification model is trained. Subsequently temporal activity context (AC) from other players on the court is added for a second, extended stage of classification.
\\\textbf{Preprocessing.}
Prior to activity recognition preprocessing is required. First, the videos are calibrated to obtain a homography projection from image coordinates $\boldsymbol{x}=(x,y)$ to real-world court coordinates $\boldsymbol{\tilde{x}}=(\tilde{x}, \tilde{y})$. Approximately 8k video frames were manually annotated by scaled bounding boxes associated with corresponding activity classes. Based on those annotations we train our HOG/HOF based classifiers, as described later in more detail, and learn two color models for player segmentation, further used for automatic player detection and generation of SC and AC descriptors, as described later. The team specific color model is learnt from front team patches while the background specific color model is learnt from rear team patches as well as all other background by training Gaussian Mixture Models (GMM). This results in probabilities for pixels belonging to foreground ($P_{fg}$) and background ($P_{bg}$). For any pixel $\boldsymbol{x}$ a color similarity measurement $M_{dyn}$ describing non-static objects (players, referees, ball, moving net, ...) is then calculated from absolute differences between a median filtered background model $BG$ and the current frame $F$ at time $i$.
%Probabilities $P_{dyn}$ for all non-static objects (players and shadows, referees, ball, moving net,...) for any pixel $\boldsymbol{x}=(x,y)$ are then calculated from absolute differences between a median filtered background model $BG$ and the current frame $F$ at time $i$.

\begin{equation}
M_{dyn}(\boldsymbol{x})=\max\limits_{c \in \{r,g,b\}} |BG^c(\boldsymbol{x})-F_i^c(\boldsymbol{x})|
%P_{dyn}(\boldsymbol{x})=P(\boldsymbol{x}|BG,F_i)=\max\limits_{c \in \{r,g,b\}} |BG^c(\boldsymbol{x})-F_i^c(\boldsymbol{x})|
\label{eq:Mdyn}
\end{equation}
Together with the fore- and background probabilities, $M_{dyn}$ is incorporated as prior information into a bayesian-like framework yielding the posterior probability $P_{player}$ for the front team players, needed for the SC descriptor and the AC descriptor where player localization estimation is done on segmented foreground regions. The estimation results are quite good, for an offset of 15cm left or right, 15cm back and 30cm forward, the accuracy remains at 93.25\% on average.

\begin{equation}
P_{player}(\boldsymbol{x}) = \frac{P_{fg}(\boldsymbol{x})\:M_{dyn}(\boldsymbol{x})}{P_{fg}(\boldsymbol{x})+P_{bg}(\boldsymbol{x})}
\label{eq:Pplayer}
\end{equation}
\\\textbf{Feature Extraction.}
Both spatial descriptors (SC, RWPC) exploit player positions during activity execution (Fig.~\ref{fig:action_locations}).
The SC descriptor is calculated using $P_{player}$ and expresses on-field player distribution. For sampled points $\boldsymbol{\tilde{x}}$ laid out in a dense grid pattern on the court area, a filled area percentage $\Pi({\boldsymbol{\tilde{x}}})$ is calculated from the corresponding rectangular fill area $\Omega_{\boldsymbol{x}}$ in the image plane. $\Omega_{\boldsymbol{x}}$ is scaled depending on the transformed position $\boldsymbol{\tilde{x}}$. %, derived from $\boldsymbol{x}$ via the homography projection.
%$\Omega_{\boldsymbol{x}}$ is spanned by upper left and lower right corner points $\boldsymbol{x_{ul}}=(x_{start}, y_{start})$ and $\boldsymbol{x_{lr}}=(x_{end}, y_{end})$ and 
$\Pi$ is obtained by summing up all foreground player probabilities within $\Omega_{\boldsymbol{x}}$ and normalizing the result with respect to the size of the area, thus expressing the appearance probability $A_p$ of a player positioned at position $\boldsymbol{\tilde{x}}$ (Fig.~\ref{fig:context}).

\begin{equation}
A_p(\boldsymbol{\tilde{x}}) = \Pi(\boldsymbol{\tilde{x}}) = \Pi(\Omega_{\boldsymbol{x}}) = \frac{\sum\limits_{\boldsymbol{{x}} \in \Omega_{\boldsymbol{x}}}{P_{player}(\boldsymbol{{x}})}}{|\Omega_{\boldsymbol{x}}|}
%\Pi = {\sum\limits_{\boldsymbol{x} \in \Omega}{P_{player}(\boldsymbol{x})}}/{|\Omega|}
\label{eq:sc_perc_fill}
\end{equation}
For dimension reduction after dense sampling, the descriptor is binned in x and y direction into $b_x$ times $b_y$ cells corresponding to court areas $\Lambda_{i_x, i_y}$ of 0.5 to 1 meters extent. The indices $i_x$ and $i_y$ are calculated from the transformed court positions $\boldsymbol{\tilde{x}}$ (Equ.~\ref{eq:binning}).

\begin{equation}
i_x\in\{1,\ldots,b_x\},\:\:\:\:i_y\in\{1,\ldots,b_y\},\:\:\:\:i_x = \lfloor \tilde{x}*b_x \rfloor,\:\:\:\:i_y = \lfloor \tilde{y}*b_y \rfloor
\label{eq:binning}
\end{equation}
\begin{equation}
%SC(i_x, i_y, \boldsymbol{\tilde{x}}) = P(player|\boldsymbol{\tilde{x}}) = \Pi(\boldsymbol{\tilde{x}}) = \Pi(\Omega_{\boldsymbol{x}})
SC(i_x, i_y) = \frac{\sum_{\tilde{x} \in \Lambda_{i_x, i_y}}{A_p(\boldsymbol{\tilde{x}})}}{|\Lambda_{i_x,i_y}|}
\label{eq:}
\end{equation}
The RWPC descriptor represents normalized two-dimensional real world position coordinates $\boldsymbol{\tilde{x}}$ as projections of player positions on the court plane. While SC and RWPC features contain spatial information, HOG and HOF retrieve shape and motion features. 
\\\textbf{Player-centered Activity Recognition.} The above mentioned features are calculated on the annotated bounding boxes to train a SVM. The output of this classifier are single frame player activity classifications. This means, that neither information about the same player at a time before or after this moment, nor information about the other players activities is incorporated.
\\\textbf{Activity Context for Player Activity Recognition.} The AC descriptor gathers information about simultaneous players activities over time. At each player position, estimated via probabilities $P_{player}$, the above trained SVM classifiers for all activities $a \in A$ are evaluated. Similar to the SC descriptor the player positions are binned within areas $\Lambda_{i_x,i_y}$, holding the average scores of the activity classifications $P_a$ for each activity over a time span of $k$ frames prior to the evaluated frame.
%The scores for each activity classification are collected by the AC descriptor (again in world coordinates) into spatial bins over different time spans. The AC descriptor matrix is of size $bx \times by \times c$, where $bx$ and $by$ denote the number of partitions in vertical/horizontal direction and can be seen as subdivision centers on the court while $c$ is the number of classes.
%Each bin is describing the average score for each activity within the chosen time span of $k$ previous frames (Fig.~\ref{fig:ac_descriptor}).

\begin{equation}
%P_{a^c} = P(a^c|\boldsymbol{\tilde{x}}) \:\:\:\:\:\:\:\:\:\:\:\:\:\:\:\:AC(bx, by, c) = {\sum\limits_{n=1}^k{P_{a_n^c}}}\big/{k}
%a \in \{A\}\:\:\:\:\:\:\:\:\:\:\:\:\:
AC(i_x, i_y, a) = \frac{\sum\limits_{n=1}^k{P_{a}^n(a|\Lambda_{i_x,i_y})}}{k}
%P_{a} = P(a|i_x,i_y) \:\:\:\:\:\:\:\:\:\:\:\:\:AC(i_x, i_y, a) = \frac{\sum\limits_{n=1}^k{P_{a}^n}}{k}
%P_{a^c} = P(a^c|\boldsymbol{\tilde{x}}) \:\:\:\:\:\:\:\:\:\:\:\:\:\:\:\:AC(bx, by, c) = {\sum\limits_{n=1}^k{P_{a_n^c}}}\bigg/{k}
%P_{a^c} = P(a^c|\boldsymbol{\tilde{x}}) \:\:\:\:\:\:\:\:\:\:\:\:\:\:\:\:AC(bx, by, c) = {\sum\limits_{n=1}^k{P_{a_n^c}}}\Bigg/{k}
%P_{a^c} = P(a^c|\boldsymbol{\tilde{x}}) \:\:\:\:\:\:\:\:\:\:\:\:\:\:\:\:AC(bx, by, c) = \frac{\sum\limits_{n=1}^k{P_{a_n^c}}}{k}
\label{eq:ac_formal}
\end{equation}
\\\textbf{Activity Recognition.} We train our classifiers for possible feature combinations using ground truth annotated training data. For evaluation on the test data we use the Bayesian player probabilities to locate unannotated players and evaluate the trained classifier on these detections. The results are collected to build the AC descriptor and concatenated with the other features for training a new classifier.
\begin{figure}[htbp!]
  \begin{center}
			\includegraphics[width=1\textwidth]{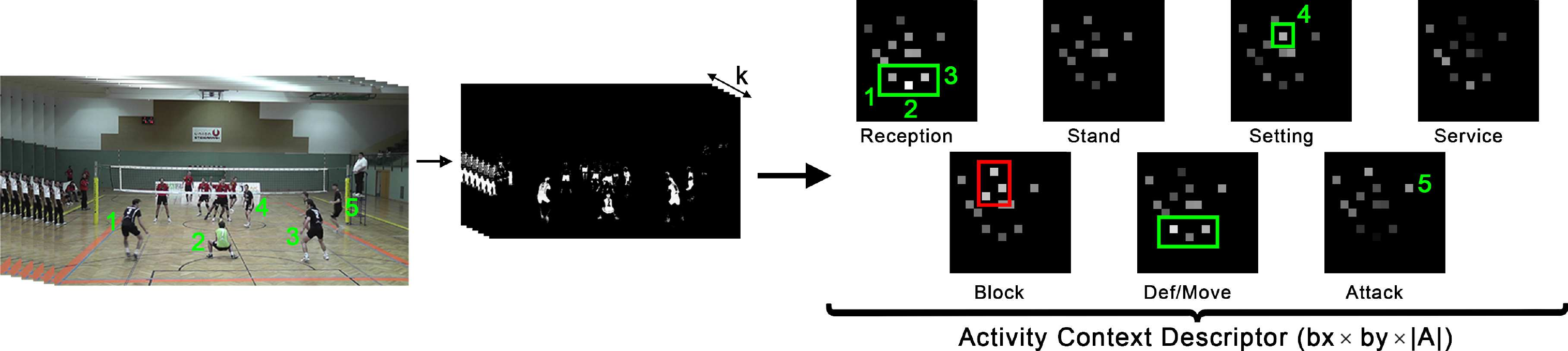}
  \end{center}
  \caption{\small\textit{Illustration of the AC descriptor: Blobs from $k$ frames are extracted, classified and the average scores are represented. For receiving players (1,2,3) the probability for the class "Setting" is low, whereas for the setter at the net (4) the probability is high. The AC descriptor is a combination of all seven class probability maps. The response for the three receiving players is evident by the high values in the "Reception" map, but also the related "Defense/Move" class shows strong responses (green rectangles). Due to the proximity to the net, the opposite players influence the "Block" map (red rectangle). Player 5 - although standing - causes a strong response in the "Attack" class, as this is a typical position for attack and the classification framework is biased by spatial information. A good example for occlusions in the video data is the sixth player, positioned behind player 3.}}
  %\caption{\small\textit{Illustration of the AC descriptor: Blobs from a series of $k$ frames are extracted, each blob is classified and the results saved in a $bx\times by$ map (averaged by $k$). For the receiving players numbered 1,2 and 3, the probability for the class "Setting" is expectedly low, whereas for the setter at the net (4) the probability is high. The AC descriptor is the combination of all seven class probability maps. The response for the three receiving players is evident by the high values in the "Reception" map, but also the related "Defense/Move" class shows strong responses (green rectangles). Due to the proximity to the net, the opposite players influence the "Block" map. This can be considered noise (red rectangle). The player marked with number 5, although standing causes a strong response in the "Attack" class, as this is a typical position for attack and the classification framework is biased by spatial information. The sixth player on the court is not marked as he is behind player 3, and is a good example for occlusions in the video data.}}
  \label{fig:ac_descriptor}
\end{figure}

\section{Experiments}
\begin{figure}[htbp!]
  \begin{center}
	  \includegraphics[width=1\textwidth]{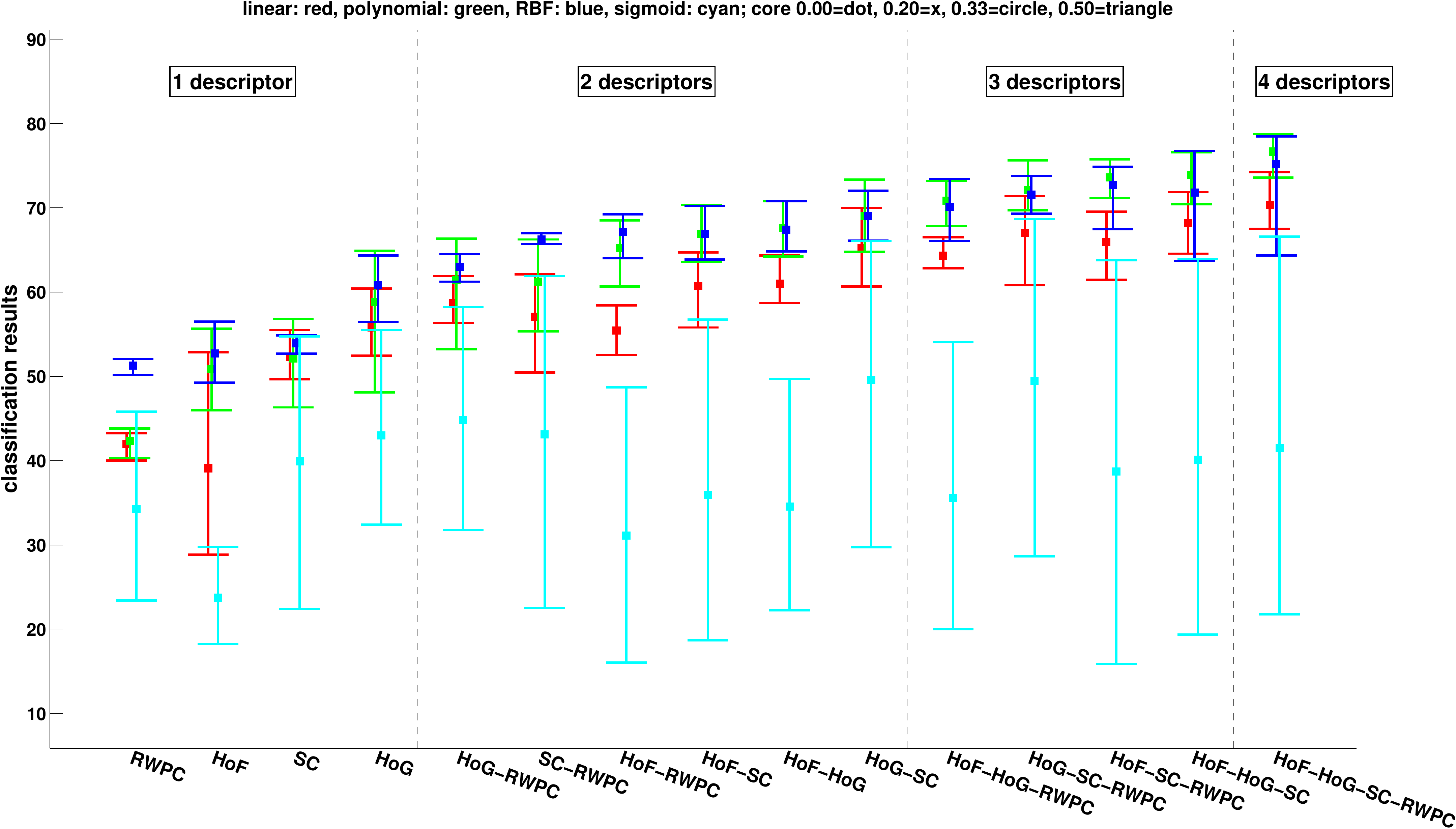}
  \end{center}
  \caption{\small\textit{Results overview: Average results for kernels and descriptor combinations over all tested parameter configurations. Polynomial (green) and RBF (blue) kernels perform better than the linear (red) kernel and the sigmoidal (cyan) kernel performs worst. Adding descriptors increases accuracy.}}
  \label{fig:allinone}
\end{figure}

\begin{figure}[htbp!]
  \begin{center}
    \subfigure[without AC]
		{
      \includegraphics[width=0.44\textwidth]{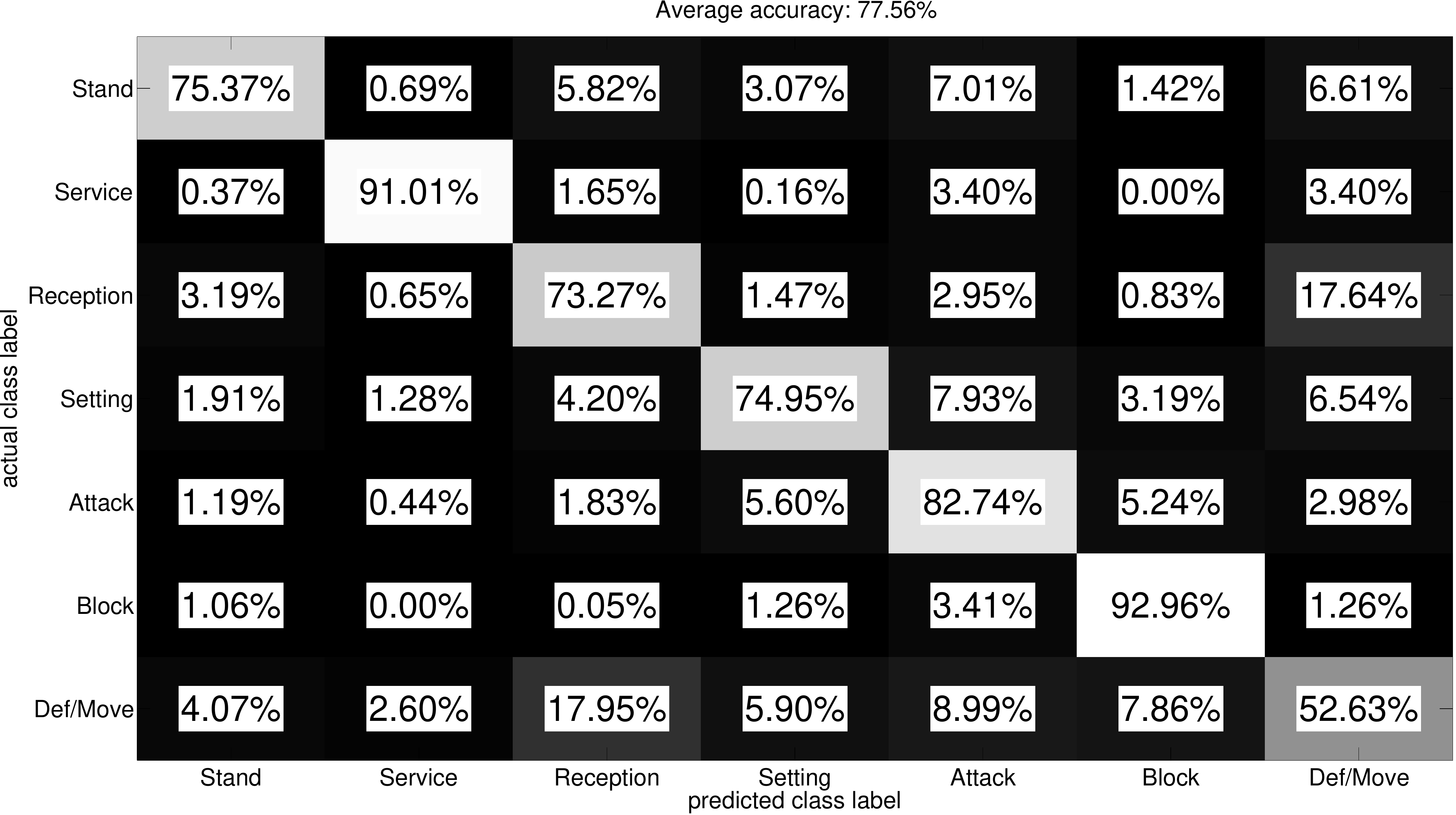}
			}
		\subfigure[with AC]
		{
      \includegraphics[width=0.44\textwidth]{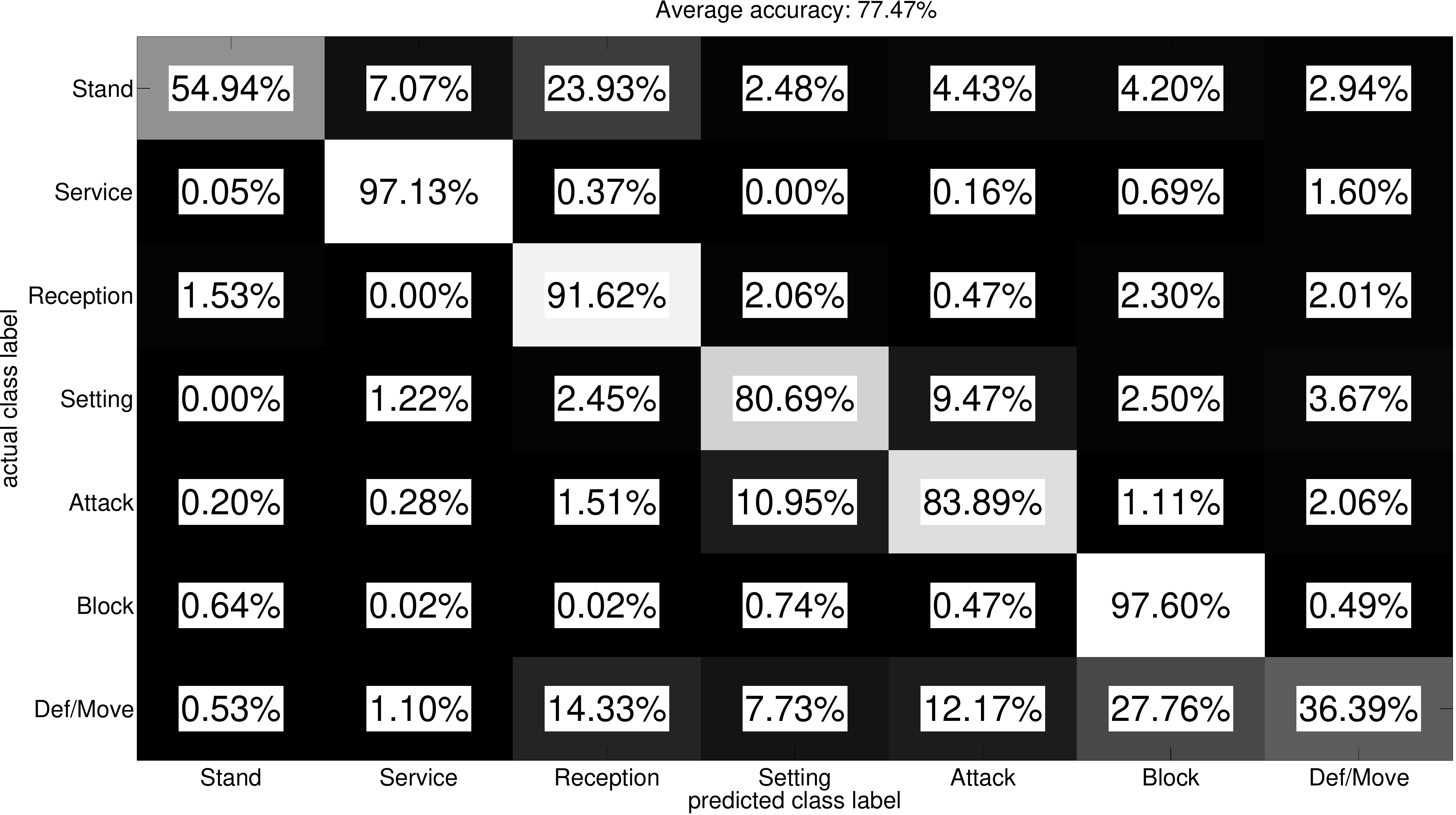}
    }			
  \end{center}
  \caption{\small\textit{Results: a) shows the results of player-centered activity recognition and b) shows results of activity context player recognition. While accuracy on the five volleyball specific activities improves by 7.20\% on average, accuracy on the two neutral classes "Stand" and "Defense/Move" decreases with added AC descriptor.}}
  \label{fig:results}
\end{figure}

\textbf{Data.} The videos were recorded from matches in the AVL\footnote{\href{http://volleynet.at/}{Austrian Volley League}} in HD resolution (1920x1080) at 25fps, compressed with the DivX codec. 6 video clips from 3 different games with a duration of approximately 2.5 hours were processed, resulting in 7973 manual annotations in seven classes. These annotations were interpolated resulting in a total of almost 36k annotations. Due to the immanent game structure, the activity occurrences differ. Also some activities like "Block" or "Stand" can be executed by multiple players simultaneously. Still, the number of activities is quite balanced. Table \ref{tab:action_quantities} shows a list of all activities and their quantities. 
%\begin{table*}[htbp]
	%\centering\scriptsize
%\begin{tabular}{|cccc|}
   %\hline
   %\multirow{2}{*}{\textbf{activity name}}& \textbf{number} & \textbf{number} & \textbf{number}\\
	  %& \textbf{of tracklets} & \textbf{of activities} & \textbf{of activities (interpolated)}\\
   %\hline
	%\hline
          %Stand & 126 & 1313 & 6067\\
          %Service & 106 & 868 & 3911\\
          %Reception & 83 & 767 & 3482\\
          %Setting & 119 & 891 & 3903\\
          %Attack & 130 & 1157 & 5233\\
          %Block & 214 & 1847 & 8332\\
          %Defense/Move & 123& 1130 & 5062\\
					%\hline
					%total & 901 & 7973 & 35990\\
   %\hline
%\end{tabular}
%\caption{\small\textit{Activity quantities: The left column shows the activity names. Tracklets are continuous player activity clips lasting about 1-2 seconds, activities denote an annotation in a video frame (every 5-10 sec) and interpolated activities denote the total number of annotations.}}
	%\label{tab:action_quantities}
%\end{table*}
\begin{table*}[htbp]
	\centering\scriptsize
\begin{tabular}{|c|ccccccc|c|}
   \hline
	& \multirow{2}{*}{Stand} & \multirow{2}{*}{Service} & \multirow{2}{*}{Reception} & \multirow{2}{*}{Setting} & \multirow{2}{*}{Attack} & \multirow{2}{*}{Block} & Defense/ & \multirow{2}{*}{total}\\
	& & & & & & & Move &\\
	\hline
	\textbf{tracklets} & 126 & 106 & 83 & 119 & 130 & 214 & 123 & 901\\
	\textbf{activities} & 1313 & 868 & 767 & 891 & 1157 & 1847 & 1130 & 7973\\
	\textbf{activities} & \multirow{2}{*}{6067} & \multirow{2}{*}{3911} & \multirow{2}{*}{3482} & \multirow{2}{*}{3903} & \multirow{2}{*}{5233} & \multirow{2}{*}{8332} & \multirow{2}{*}{5062} & \multirow{2}{*}{35990}\\
	\textbf{(interp.)} & & & & & & & &\\
   \hline
\end{tabular}
\caption{\small\textit{Activity quantities: The top row shows the activity names. Tracklets are continuous player activity clips lasting about 1-2 seconds, activities denote a manual annotation in a video frame (every 5-10 frames) and interpolated activities denote the total number of annotations.}}
	\label{tab:action_quantities}
\end{table*}
While the classes "Service", "Reception", "Setting", "Attack" and "Block" are specific volleyball activities, the two other classes "Stand" and "Defense/Move" are more general classes. The latter is a very inhomogeneous and hard class, as all activities that do not fall into any of the other categories are collected within this class. The data was partitioned into 50\% for training and testing. We plan to make the dataset available for future research.
\\\textbf{Parameters.} The following parameters were chosen for descriptors and classifiers: HOG/HOF (cell size, patch width+height, cells per block, bins), SC (horizontal/vertical grid spacing, horizontal/vertical binning), AC (number of considered frames, horizontal/vertical binning), SVM (cost parameter $c$, kernel parameter $\gamma$, kernel type (linear, sigmoid, RBF, polynomial)). Together with the possible set of descriptor combinations, we have tested a total of over 750 combinations.
\\\textbf{Results.} Results vary strongly dependent on the choice of parameters of the descriptors and SVM. We have tested all 15 descriptor combinations intensively and showed that adding descriptors improves performance (Fig.~\ref{fig:allinone}).
The best result achieved for player-centered activity recognition under any parametrization of the four descriptors HOG, HOF, RWPC and SC yields an average accuracy of 77.56\%. While for the "Defense/Move" class with 52.63\% the result might be considered rather poor, the other classes perform very well with 73.37\% to 92.96\% correctly classified activities.
\begin{figure}[htbp!]
  \begin{center}
      \includegraphics[width=.95\textwidth]{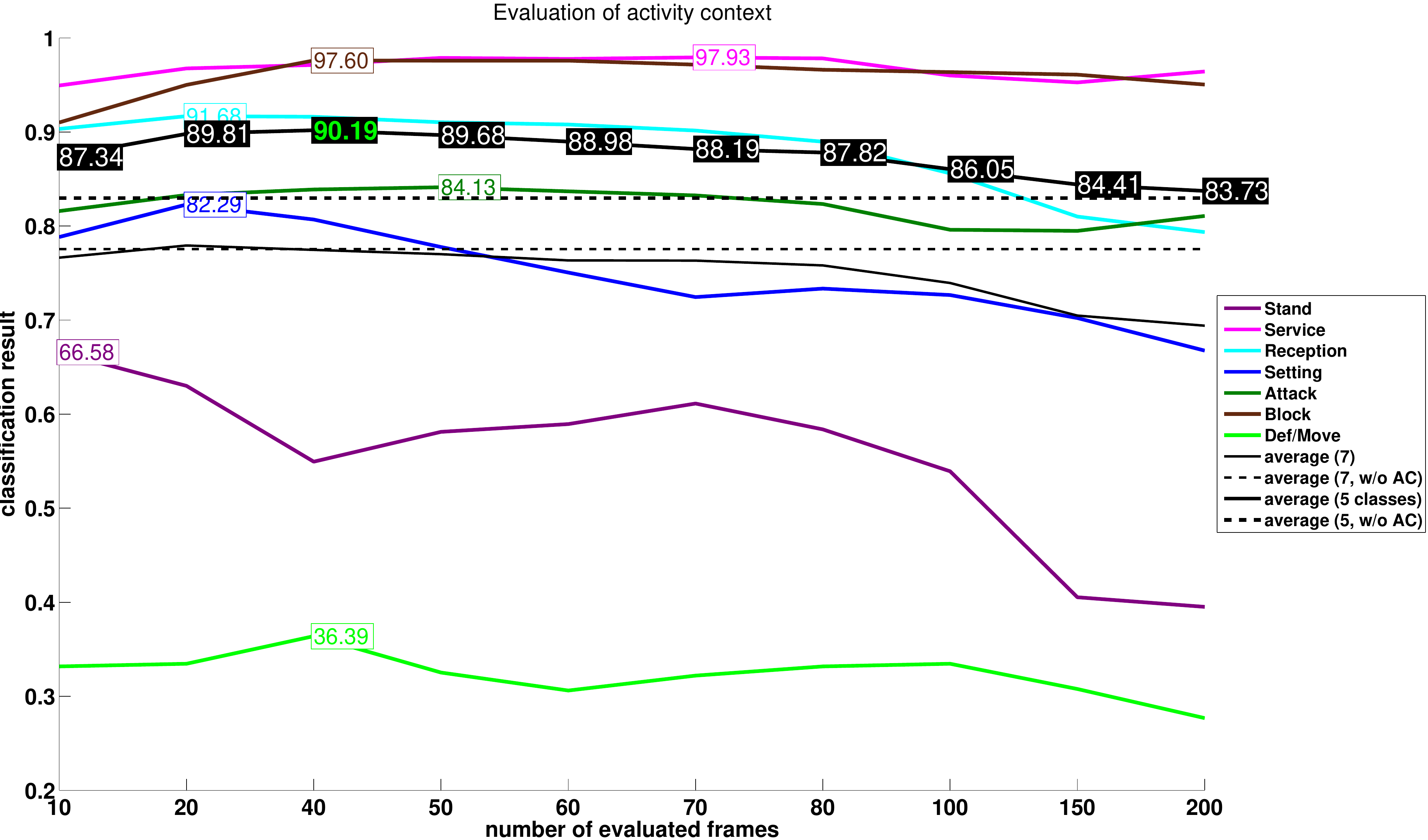}
	\end{center}
  \caption{\small\textit{Results for activity recognition with AC and differing $k$: The best result (90.19\%) on the 5 volleyball specific classes is achieved with $k=40$ frames (1,6s).}}
  \label{fig:ac_results10_k3}
\end{figure}
When adding the AC descriptor also chronological order and correlation of activities is introduced. Therefore it is not surprising, that the two general classes "Stand" and "Defense/Move" deteriorate as they occur almost randomly throughout the games and have no other specific activities occurring in temporal context. Investigating 40 frames or 1.6 seconds before the actual frame gives best average result. Fig.~\ref{fig:ac_results10_k3} shows performance in dependence of the length of the time-span included for building the AC descriptor.
For the volleyball specific activities, the results improve by 7.20\% on average and all five activities are above 80\% with top result for "Block" (97.60\%) and "Service" (97.13\%). For the "Reception" class, the similarity confusion with "Defense/Move" can be removed by the AC descriptor, improving recognition by 18.35\% to 91.62\%. The 7 class average is approximately identically to without AC, while the average on the 5 specific classes is better for any value of $k$. See Fig.~\ref{fig:results} for AC descriptor results.
\section{Conclusion}
We presented an evaluation of single player activity recognition on a new indoor volleyball dataset. Starting the classification from standard features (HOG, HOF, position) trained from manual annotations, we further incorporated activity recognition scores of automatically detected players and integrated them as contextual information by the proposed activity context (AC) descriptor. This improves the results for single player activity recognition by up to 18.35\% and 7.20\% on average on volleyball specific actions within our new dataset. This proves that contextual knowledge about simultaneously executed activities of other team members supports classification of individual player activities.
\\\textbf{Outlook.} Additional features (trajectories, velocities, 3D information, ball position, opponent activities) could further improve results. Extension to recognition of team activities or other team sports would also be interesting. The proposed method of incorporating spatial and temporal context for improved single person activity recognition should be extensible to other areas like surveillance or home care, where the observed person is also in a relationship with surrounding persons or objects.

%\input{5_conclusion}

%\small
\footnotesize
\bibliography{refs}
\end{document}